\documentclass{ar-1col-S2O}

\usepackage[comma]{natbib}
\usepackage{url}
\usepackage{custom}

\jname{Annu. Rev. Stat. Appl.}
\jvol{TBD}
\jyear{2025+}
\doi{10.1146/TBD}

\begin{document}

\markboth{Dobriban}{Statistics in Generative AI}

\title{Statistical Methods in Generative AI}

\author{Edgar Dobriban$^1$
\affil{$^1$Department of Statistics and Data Science, University of Pennsylvania, Philadelphia, PA, USA, 19104; email: dobriban@wharton.upenn.edu}}

\begin{abstract}
Generative Artificial Intelligence is emerging as an important technology, promising to be transformative in many areas. 
At the same time, generative AI techniques are based on sampling from probabilistic models, and by default, they come with no guarantees about correctness, safety, fairness, or other properties.
 Statistical methods offer a promising potential approach to improve the reliability of generative AI techniques. 
In addition, statistical methods are also promising for improving the quality and efficiency of AI evaluation, as well as for designing interventions and experiments in AI.
 In this paper, we review some of the existing work on these topics, explaining both the general statistical techniques used, as well as their applications to generative AI. We also discuss limitations and potential future directions. 
\end{abstract}

\maketitle

\tableofcontents

\section{Introduction}

     Artificial Intelligence, and more specifically, Generative AI,  is emerging as an important technology. 
     Over the past few years a number of prominent generative AI technologies have been developed and have received widespread attention; ranging from text generation via large language models 
     (ChatGPT, Claude, Llama, Gemini, DeepSeek, Qwen, etc), 
     image generation via diffusion models (Dall-E, Stable Diffusion, etc), to scientific generative AI techniques used for protein generation \citep[e.g.,][etc]{watson2023novo}, DNA sequence editing \citep[e.g.,][etc]{ruffolo2025design}, among others.

     Such methods have been quickly adopted by end users and institutions, both via direct usage, as well as integrated in other tools such as code assistants and web search agents.
     The scientific community has shown significant interest in using generative AI models, achieving a number of breakthrough results \citep[see e.g.,][etc]{davies2021advancing,hayes2025simulating}, culminating in a 2024 Nobel Prize in Chemistry awarded in part for work with a significant component in protein structure design and generation \citep{nobel2024chem}. 

Yet, the adoption of generative AI (GenAI) methods more generally is hindered by their lack of reliability \citep[see e.g.,][etc]{farquhar2024detecting,strauss2025real,manduchi2025on}. 
At their core, these methods rely on sampling from probability distributions over complex spaces that are learned from huge datasets.
 At the outset, GenAI does not provide any guarantees about correctness, safety, or any other desired criteria. 
While the performance and reliability of GenAI models is increasing steadily, so far, issues around reliability have not been successfully eliminated.

\begin{marginnote}
\entry{Generative AI}{The construction of probabilistic models over large semantic spaces (text, images, etc.) that allows sampling from these models, given certain inputs.}    
\end{marginnote}

Statistical methods offer potential opportunities to improve the reliability of GenAI systems. 
In this paper, we review several examples, highlighting statistical methods with proven or potential applications in generative AI.
 We focus on four topics: 
 improving and changing the behavior of systems,
 diagnostics and uncertainty quantification,
 AI evaluation,
 as well as interventions and experiment design.
We highlight here that the approaches we discuss are as of now mainly in the research phase, and they are usually not yet deployed in mainstream generative AI products. Their eventual usefulness remains to be determined.
 
\subsection{About This Review}
Generative AI models are commonly studied separately, for each specific modality that they pertain to (text, images, video, etc),  
or based on the underlying technology (diffusion models, large language models or LLMs, etc). 
There are already a few reviews with significant coverage of statistics related to these topics individually, including
\cite{chen2024overview,zhang2025generalization} for diffusion models,
\cite{suh2024survey} for deep learning more generally, but also touching on generative models, 
and
\cite{ji2025overview} for
language models.
 
Our focus is different, rendering our work largely non-overlapping with the above works. We focus on techniques that are applicable to all generative AI models, regardless of their modality. 
Moreover,  
with a few exceptions, do not focus on
statistical methods that are applicable to generative AI only when the tasks of interest are essentially simple classification/regression tasks (e.g., multiple-choice question-answering with LLMs). 
For these, there are already numerous useful references.
 
 We also focus specifically on statistical methodology for AI and omit discussion about  statistical theory of generative AI, 
as well as statistics-adjacent methods that primarily leverage optimization or other techniques.
Further, we omit certain topics, such as watermarking and ranking from human preferences, which have already been discussed in detail in the above  works.
Due to space limitations, we mainly consider simple methods that have clear theoretical motivation and often provable guarantees.
Moreover, we also omit discussion of how generative AI models can be used to improve statistical analysis, see e.g., \cite{bashari2025synthetic} for a representative example.

{\bf Target audience.}
Our target audience includes statisticians eager to see how their expertise can drive impact in generative AI, AI researchers interested in how statistical methods can strengthen their tools, and scientists looking to better understand this emerging area. 
 For this reason, our paper aims to be largely self-contained, with prerequisites that include knowledge of introductory undergraduate-level probability and statistics, and a basic familiarity with AI at the advanced undergraduate level.


\subsection{What is Generative AI?}

\begin{table}
\tabcolsep7.5pt
\caption{Representative types of generative AI models and their input and output spaces.}
\label{tab:predictive_models}
\begin{center}
\begin{tabular}{@{}l|c|c@{}}
\hline
Generative AI Model $\hat{p}$ & Input Space $\mathcal{X}$ & Output Space $\mathcal{Y}$ \\
\hline
Language Models & Text & Text \\
Diffusion Models & Text & Images \\
Multimodal Language Models & Images, text & Images, text, sound, video \\
Protein Structure Generation & Amino acid sequence & 3D structure \\
\hline
\end{tabular}
\end{center}
\end{table}
Generative AI usually refers to the use of generative models, which are learned probability distributions one can sample from.
Concretely, consider an input space 
$\mX$ (e.g., images, text, documents, their combinations, etc., represented in an appropriate way)
and an output space 
$\mY$ (similarly, this could be images, text, audio, video, etc).
See Table \ref{tab:predictive_models} for some examples.

Formally speaking, this includes as a special case standard statistical machine learning problems such as classification (when $\mY$ consists of the classes) and regression (when $\mY=\R$, for instance).
However, the cases of interest in generative AI are usually 
high-dimensional spaces $\mY$ representing objects 
that are 
semantically meaningful to humans, such as text---viewed as a sequence of symbols $(x_1, \ldots, x_k) \in V^k$ for a finite set $V$ of symbols---or images, viewed as tensors representing pixels.
\begin{marginnote}
\entry{Generative Model}{A generative model $\hat p$ provides a way to 
sample an output $Y\sim \hat p(\cdot\mid x)$ from the conditional distribution of $\hat p$ given any input $x\in \mathcal{X}$.}    
\end{marginnote}

\begin{figure}
  \centering
  \includegraphics[width=4in]{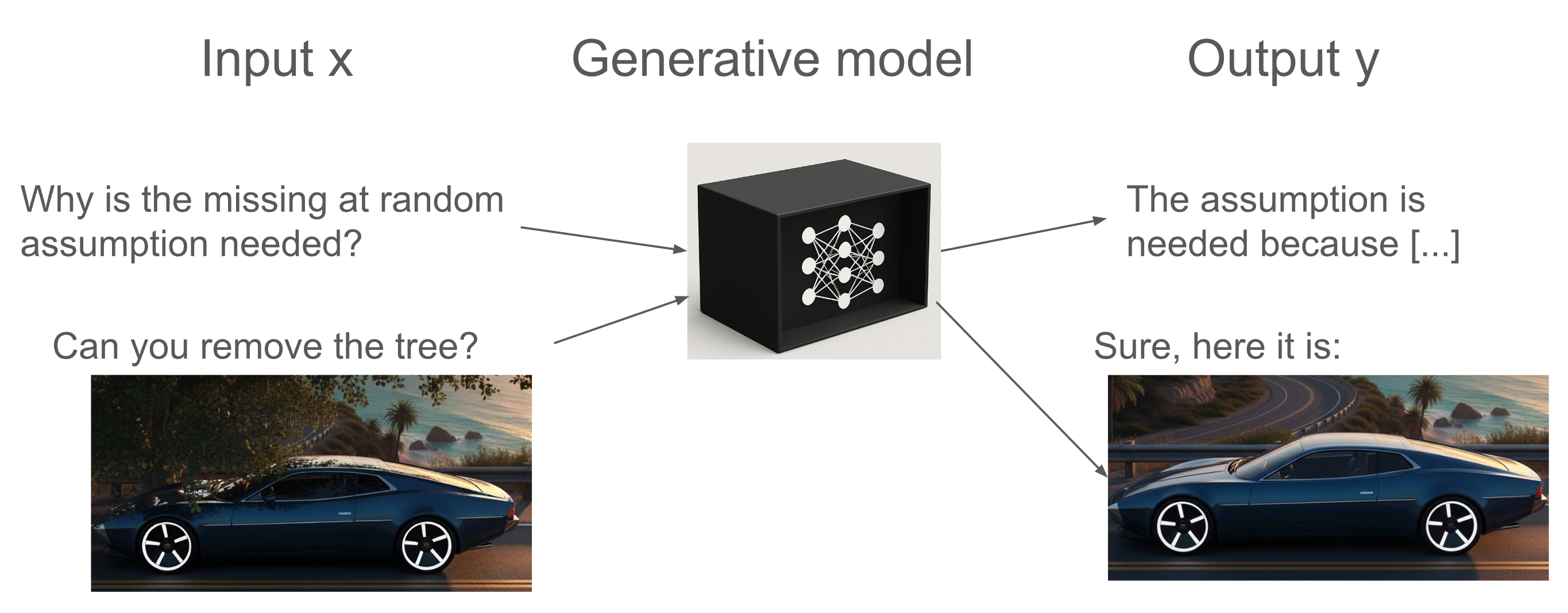}
  \caption{General workflow of a generative model: inputs (e.g., text prompts, images) are processed through a black-box model to produce outputs.}
  \label{fig:workflow}
\end{figure}

Generative AI models are often designed for interaction with humans. 
A simple protocol is as follows:
The user inputs a specific $x\in \mX$, for instance, a text prompt such as “How can I fix a broken lamp?".
 Then, the generative model $\hat p$ provides a way to 
draw a sample $Y\sim \hat p(\cdot\mid x)$ from the conditional distribution $\hat p$ given $x$; for instance, a textual response by the language model such as “To fix a broken lamp, you need to [...]".
This is then returned to the user.
See Figure \ref{fig:workflow} for an illustration.
The ability to provide a user input corresponds to being able to sample conditionally from the generative model.
This crucially unlocks a huge range of applications, by being able to be responsive to the specific needs of the user.\footnote{This aspect is especially striking when compared to the lack of this capability in previous versions of generative models, such as Generative Adversarial Networks (GANs) or Variational Autoencoders (VAEs). It is one of the crucial technological advances that explain the utility of generative AI models, see e.g., \cite{rombach2022high,zhang2023adding,parmar2023zero}, etc.}
The interaction can also continue. For simplicity, we will mostly restrict our discussion to one round of interaction.


\subsection{How is a Generative Model Learned?}
The GenAI model $\hat p$ is usually obtained by empirical loss minimization, in a manner that is conceptually similar to that used in most standard statistical modeling and machine learning. 
This is performed by running 
an algorithm---often a stochastic gradient descent-based method or a variant---aiming to minimize a loss function over a large function class using a {massive data set}.

    For instance, for language models, the training data consists of text represented as a collection of sequences 
    $x = (x_1, x_2, \ldots, x_{k})$, where 
    for a finite set $V$ usually referred to as a vocabulary, each $x_j \in V$, $j\le k$. 
    The length $k$ of the strings can vary, up to a so-called context length $L$.
    Instead of viewing text as a sequence of letters, usually, text is encoded in tokens which are adjacent groups of letters  that can offer more efficiency in the modeling process.
    For instance, “encoded" might consist of the tokens “en+code+d".

    The loss used is often the negative log-likelihood 
    $\theta \mapsto -\sum_{x\in \mathcal{D}} \log p_\theta(x)$. 
    The  
    function class $\theta\mapsto p_\theta$ usually consists of huge neural nets parametrized in very special ways, with up to hundreds of billions of parameters. 
    The dataset used for training consists of text data crawled from  the internet, enriched with high information content (Wikipedia, arXiv), and other sources such as books.
    Typical costs for training powerful Generative AI models can start from millions of US dollars, which means that only organizations with significant financial resources can perform the initial training.

\subsection{Access Mode to the Generative Model}

An important consideration is the mode of access that we have to the generative model of interest. 
     At the time of writing, the most powerful GenAI models are closed-source and run by commercial providers on their own cluster infrastructure,  accessible only through querying. 
    This leads to a \emph{black box} mode of access, meaning that for any given input $x$, we can only observe the output $Y$, but not any internal components of the generation process $\hat p$. Sometimes some additional information is provided in a \emph{gray box} access mode; for instance, the probability $\hat p(Y|x)$ may also be returned.
\begin{marginnote}
\entry{Black Box Access}{An access model where we can only observe the output of a GenAI model, and not its internal workings.}    
\end{marginnote}    

     Open-source or open-weights GenAI models may be run on local machines depending on the available hardware.\footnote{They typically require powerful graphics processing units (GPUs) to be run with a reasonable speed.} In such cases, it is possible to inspect the internal workings of the models. However, since generative models tend to be highly complicated neural networks, using the internal information is challenging.
     Therefore, to maintain generality, we will usually focus on methods applicable to black box GenAI models. In a few cases, we will also discuss methods that require gray or white box access.

\section{Statistical Methods in Generative AI}

Our goal is to
discuss a few emerging areas of research where statistical methods or ideas can be used in generative AI. 
A key starting point is that AI systems can be wrong. 
They can make any type of mistake, and they have no guarantees by default about correctness, content, logical consistency, safety, etc.\footnote{At the moment, it is only possible to rigorously understand and analyze individual components of GenAI models in isolation, see e.g.,  \cite{noarov2025foundations} for an example of analyzing the final decoding step in language models.}
This stems intrinsically from their structure as sampling methods.\footnote{It is often possible to ensure that the generation process is deterministic; for instance, in large language models, one can set the temperature parameter to zero. However, the resulting deterministic generative models still inherit the lack of intrinsic correctness due to the black box nature of the original model.} 

While there are a variety of engineering approaches to improve reliability, such as endowing the AI models with external tools, such as calculators, web search, or access to a computer where they can run programs, the use of these tools is in turn orchestrated by a sampling-based generative AI model, which can still have reliability problems.  Moreover, while there are constrained sampling methods that aim to ensure certain basic formatting and correctness criteria, their current scope is limited; for example, at the moment they cannot ensure logical correctness.

 For these reasons, statistical methods that aim to improve the behavior of generative models---sometimes with provable guarantees---are particularly significant; we begin our discussion with this topic. 
Crucially, to have an impact in this area, statistical methods must directly align with AI practice and goals; endowing practically useful AI-enhancement methods with desirable guarantees. 
To put it another way, statistical methods act as simple tunable wrappers that can be calibrated to meet explicit error budgets with finite sample guarantees.

\subsection{Improving and Changing Behavior}
\label{sec-imp}

To improve the performance of a generative AI model,
 there are numerous of standard approaches
 relying on variants of standard training (e.g., supervised fine-tuning for LLMs). 
Once these have been exhausted, there is room for alternative techniques that change the behavior of the generative model in a non-standard way that can conceivably improve certain accuracy metrics, for instance, by returning a trimmed version of the input from which false claims have been deleted \citep[see e.g.,][etc]{mohri2024language}.
These techniques can be roughly categorized into changing 
    (a) the output $y$, 
    (b) the input $x$, or 
    (c) the internal workings on the generative model $\hat p$. 

Moreover, many of these techniques require a degree of hyperparameter tuning; for instance, determining how much to trim the outputs.
 This process of tuning can sometimes be endowed with statistical correctness guarantees (see Table \ref{beh}), and so this is the first topic we review in this work.

\begin{table}
\tabcolsep=5.5pt
\caption{Types of methods that change the behavior of generative AI systems; most of them endowed with statistical guarantees. Some methods belong to multiple categories.}
\label{beh}
\begin{center}
\begin{tabular}{@{}c|c|p{9cm}@{}}
\hline
\textbf{Technique} & \textbf{Type} & \textbf{Examples} \\
\hline
\multirow{8}{*}{\shortstack{Change\\output}} 
    & \multirow{2}{*}{\shortstack{Additional\\output type}} 
        & Highlight parts of output \citep{sun2022investigating,vasconelos2025generation} \\
\cline{3-3}
& 
        & Abstain from generation when a risk score is high \citep{farquhar2024detecting,yadkori2024mitigating} \\
\cline{3-3}
    & 
        & Add “Everything Else” as a possible answer \citep{noorani2025conformal} \\
\cline{2-3}
    & \multirow{3}{*}{\shortstack{Set of\\outputs}} 
        & Construct prediction interval for each output coordinate \citep{horwitz2022conffusion,teneggi2023trust} \\
\cline{3-3}
    & 
        & Generate set of outputs \citep{quach2024conformal,gui2024conformal,nag2025conformal} \\
\cline{2-3}
    & \multirow{2}{*}{\shortstack{Trimmed\\output}} 
        & Delete parts of output until correctness is achieved \citep{khakhar2023pac,mohri2024language} \\
\cline{3-3}
    & 
        & Find small parent set of possible outputs in a directed acyclic graph \citep{zhang2024conformal} \\
\cline{2-3}
    & \multirow{1}{*}{\shortstack{Regenerated\\output}} 
   &
        Reformulate output until it is appropriately correct and specific \citep{jiang2025conformal} \\
\cline{2-3}
    & \multirow{1}{*}{\shortstack{Task-specific\\output}} 
        & Train model to improve performance in downstream task \citep{band2024linguistic}
      \\
\cline{3-3}
    & 
        & Construct prediction intervals for latent variables of a generated output  \citep{sankaranarayanan2022semantic}
      \\
\cline{3-3}
    & 
        & Interactively ask questions that maximize the informativeness of the answers  \citep{chan2025conformal}
      \\      
\hline
\multirow{1}{*}{\shortstack{Change\\input}} 
    & \multirow{1}{*}{\shortstack{Set of\\inputs}} 
        & Retrieve sets of documents in RAG \citep{li2024traq}\\
\cline{3-3}
       && 
        Select prompts that control risk \citep{zollo2024prompt} \\
\hline
\multirow{4}{*}{\shortstack{Change\\other\\algorithm\\settings}} 
    & \multirow{1}{*}{\shortstack{---}} 
        & Accelerate generation by early exit \citep{schuster-etal-2021-consistent,Schuster2022confident,jazbec2024fast} \\
\cline{3-3}
    & 
        & Reduce ambiguity by seeking additional input \citep{ren2023robots,ren2024explore} \\
\cline{3-3}
    & 
        & Control a ``size" component of the sampling mechanism \citep{ravfogel-etal-2023-conformal,deutschmann2024conformal,ulmer-etal-2024-non} \\
\cline{3-3}
    & 
        & Switch between models when risk score is high \citep{overman2025conformal} \\
\hline
\end{tabular}
\end{center}
\end{table}

\subsubsection{An example: Controlling the probability of refusal/abstention}
\label{refu}
To get a sense of the types of problems that can be solved, as well as the types of statistical methods that are used, we will explain one specific example in some detail.
We will consider the example of abstaining from generation when a risk score is high \citep[see e.g.,][etc]{farquhar2024detecting,yadkori2024mitigating}.

         Consider a given loss function\footnote{In the literature cited above, this is sometimes called a risk score, but we will not use that term, in order to avoid a conflict of terminology with the classical notion of risk---namely, expected loss---from statistical decision theory.} 
        \(
            \ell : \mathcal{X} \times \mathcal{Y} \to \mathbb{R}.
        \) This could measure the quality or safety of an input-output pair. There are many examples, including the negative log likelihood $\ell(x,y) = -\log \hat p(y|x)$ specified by the generative model itself, or the negative of a pre-trained reward function (measuring for instance safety), etc. 
        The loss could depend on both $x$ and $y$, or only on one of the two.
         If the loss only depends on the input $x$, it can capture either input ambiguity, or the dispersion in outputs generated by the model \citep{lin2023generating}; or some combination thereof.
         
         \begin{marginnote}
\entry{Refusal/Abstention}{When a generative AI model does not return an output. Can be useful for improving safety.}    
\end{marginnote}  
         To improve user experience, a strategy is to  refuse/refrain/abstain from answering when the loss is high.
         Specifically, we want to find a threshold $\tau$ such that  when $\ell(x,Y) > \tau$ we should instead return a special message like \texttt{``Sorry I cannot answer.''}, where $Y \sim \hat{p}(\cdot|x)$ is generated by the model $\hat p$.
         There is a trade-off: decreasing the threshold will ensure that only higher quality---lower loss---generations/answers are returned, but higher refusal also hampers utility to users.
         
         The threshold $\tau$ can be set by standard hyperparameter tuning, by checking the loss values and abstention rates on a dataset.
         However, there is also a statistical approach, which can provide provable guarantees on the behavior of the system under certain conditions. This approach is based on predictive inference/conformal prediction \citep{vovk2005algorithmic}, and the ideas date back to work on tolerance regions \citep[e.g.,][etc]{wilks1941determination,wald1943extension}. 
         \begin{marginnote}
\entry{Predictive Inference}{The goal of endowing the outputs of predictive models---including GenAI---with statistical guarantees.}    
\end{marginnote}

         The statistical approach aims to guarantee generalization to a distribution $D$ of prompts.
         The goal is then to control the abstention probability 
         over the distribution $D$, which can be written as
        \(
            \Pr_{X \sim D,\, Y \sim \hat{p}(\cdot|X)}\!\big(\ell(X,Y) > \tau \big).
        \)
         We do not fully know the distribution $D$, because it represents the behavior of future users. However, we assume that we have a \emph{calibration dataset}---also referred to as a validation or hold-out dataset---$D_n = \{X_1, \dots, X_n\}$ of prompts which we view as an i.i.d.~sample from $D$. This is collected based on user interactions that are representative of the distribution, and we assume that they have not been used for model training.
        \begin{marginnote}
\entry{Calibration dataset}{Given a trained GenAI model, a separate dataset used to endow the model with various statistical properties.} 
\end{marginnote}

         Then, we aim to construct an estimated threshold  $\hat{\tau}=\hat{\tau}(D_n)$ using the calibration dataset $D_n$ such that the abstention probability\footnote{Notice that this probability now also includes the randomness over $D_n$.} is controlled at a user-specified level $\alpha > 0$, i.e.,
        \(
            \Pr_{X \sim D,\, Y \sim \hat{p}(\cdot|X),D_n}\!\big(\ell(X,Y) > \hat{\tau}(D_n)\big) \le \alpha.
        \)
        
         The key observation is the following: suppose we generate responses $Y_i \sim \hat{p}(\cdot|X_i)$ for each of our inputs  $i = 1, \dots, n$ from the calibration dataset. 
         Then the values $\ell_i:= \ell(X_i, Y_i)$, $i=1, \ldots, n$ are i.i.d.~random variables with the same distribution as the test loss $\ell(X,Y)$ where $X \sim D$ is a test data point and $Y \sim \hat{p}(\cdot|X)$ is a corresponding outcome sampled from the generative model.
        Of course, the distribution of these loss values is in general still unknown, because it depends on the unknown target distribution $D$. 
        
        {\bf Exchangeability.} However, since the $\ell_i$ are i.i.d., conditional on the set (or multiset) of their values 
        $S_{n+1}=\{\ell_1, \ldots, \ell_n, \ell(X,Y)\}$, their ordering is uniform given $S_{n+1}$. This corresponds to exchangeability, and it is their only property used here. 
        \begin{marginnote}
\entry{Exchangeability}{Informally, the property that a sequence of random variables is equally likely to be presented in any order.} 
\end{marginnote}

         Therefore, assuming for simplicity of exposition that there are no ties,\footnote{The extension to the case of ties is not hard, but it can require some care; see for instance \cite{vovk2005algorithmic,angelopoulos2023conformal}.} the rank of $\ell(X,Y)$ among 
        $\ell_1, \ldots, \ell_n, \ell(X,Y)$ is distributed uniformly over $\{1, \ldots, n+1\}$, conditional on $S_{n+1}$. 
         Now, for any $\beta \in [0,1]$, let $Q_\beta$ be the $\beta$-th quantile of $\{\ell_1, \ldots, \ell_n\}$, 
        namely 
        \(
            Q_\beta = \inf \{ t : \# \{ i : \ell_i \le t \} \ge \beta n \}.
        \)
         We have that 
        \(
            \ell(X,Y) \ge Q_{(1-\alpha)(1+1/n)}
        \)
        if and only if the rank of $\ell(X,Y)$ among $\{\ell_1, \ldots, \ell_n, \ell(X,Y)\}$ is at most $\lfloor\alpha(n+1)\rfloor$. 

{\bf Consequences of exchangeability.}
        By exchangeability, 
        this occurs with probability at most $\lfloor\alpha(n+1)\rfloor/(n+1) \le \alpha$, conditional on $S_{n+1}$. 
         Hence, if we choose 
        \(
            \hat{\tau}(D_n) := Q_{(1-\alpha)(1+1/n)},
        \)
        we find 
        \(
            \Pr\big(\ell(X,Y) \ge \hat{\tau}(D_n)\big|S_{n+1}) \le \alpha,
        \)
        for any $S_{n+1}$.
        Since this holds for any set $S_{n+1}$, it also holds unconditionally, i.e., \(
            \Pr\big(\ell(X,Y) \ge \hat{\tau}(D_n)\big) \le \alpha,
        \)
        as desired. 

The above argument explains how, by choosing a threshold for abstention equal to a particular quantile of the calibration losses, 
we can control the abstention rate. 
All that is needed is that the test loss is exchangeable with the calibration losses. 

\subsubsection{An overview of applications and techniques}

The above discussion is quite representative of 
a variety of methods
designed to improve the behavior of generative AI models.
Some of the common elements include: (1) introduction of
    a loss function; (2) introduction of a small number of tunable hyperparameters; (3) formulation of a desired goal in terms of expectations of the losses and probabilistic properties, and (4) using distribution-free or only weakly distributionally dependent probabilistic tools---such as  the distribution of order statistics or concentration inequalities---to ensure the desired goal.
 We refer to the works cited in Table \ref{beh} for details.
    
    For instance, one approach proposes to delete claims from the output $Y$ of a large language model until correctness is reached  \citep{khakhar2023pac,mohri2024language}. 
    This approach defines a deletion operator $\Delta$, typically implemented by another large language model, and a sequence $Y^{(k)} = \Delta(Y^{(k-1)})$, $k\ge1$, starting with  $Y^{(0)}=Y$. The loss function $\ell$ is defined based on whether $Y^{(k)}$ has any claim that contradicts a ground truth answer $y^*$ for the prompt $x$.
    This is also evaluated by another large language model.
    The tunable hyperparameter is the number $k$ of deletions; and the goal is to ensure correctness with probability at least $1 - \alpha$. Then the required number of deletions can be determined based on a calibration dataset, similarly to above. 

Many of the methods discussed above rely on some form of non-parametric statistics, distribution-free predictive inference, conformal prediction, and variants.
The idea of distribution-free prediction sets dates back at least to the pioneering works of \cite{wilks1941determination}, \cite{wald1943extension}, etc.
Distribution-free inference has been extensively studied in recent works \citep[see, e.g.,][etc]{saunders1999transduction,vovk1999machine,papadopoulos2002inductive,vovk2005algorithmic,vovk2012conditional, lei2013distribution,lei2014distribution,lei2018distribution,guan2023localized, romano2020classification,dobriban2025symmpi}. 
Predictive inference methods
have been developed under various assumptions
\citep[see, e.g.,][]{geisser2017predictive,bates2021distribution,park2021pac,park2022pac,sesia2022conformal,qiu2023prediction,li2022pac,kaur2022idecode,si2024pac,lee2024simultaneous}.
Overviews of the field are provided by \cite{vovk2005algorithmic, shafer2008tutorial}, and \cite{angelopoulos2023conformal}. 

\subsection{Diagnostics
and Uncertainty Quantification}

When AI systems encounter problems, one should of course aim to improve the behavior of the AI system.
A crucial step toward this is to precisely diagnose the problem.
A variety of approaches exist for this task, 
ranging from constructing unit tests to fine-tuning the model. 
There are also a number of methods
based on computing certain specific diagnostic scores \citep[e.g.,][etc]{farquhar2024detecting,yadkori2024mitigating,lin2023generating}.
Such diagnostics are already used in many of the methods discussed in Section \ref{sec-imp} to change or improve the generative model; for instance, if a safety score for the input is low, the model can refrain from generating an output.

In this section,
we are specifically interested in diagnostics that aim to quantify uncertainty, as these have close connections to probability and statistics.
There are a variety of interpretations of uncertainty quantification, see e.g., 
\cite{baan2023uncertainty,shorinwa2024survey,liu2025uncertainty,abbasli2025comparing,xia2025survey,he2025survey,campos2024conformal,trivedi2025need}.
 Due to space reasons, here we can only discuss a few specific approaches, see Table \ref{diag}.

\begin{table}
\caption{Types and examples of 
uncertainty quantification
methods for generative AI. }
\label{diag}
\centering
\begin{tabular}{@{}c|c|p{9cm}@{}}
\hline
\textbf{Approach} & \textbf{Type} & \textbf{Examples} \\
\hline
\multirow{1}{*}{\shortstack{Defining \\ Uncertainty}} 
&
\multirow{1}{*}{\shortstack{Epistemic \&\\Aleatoric Unc.}} 
& Define and estimate epistemic and aleatoric uncertainty through input clarification ensembling \citep{hou2024decomposing}
\\
\cline{2-3}
&\multirow{1}{*}{\shortstack{Semantic\\Uncertainty}} & Cluster outputs to capture semantic uncertainty \citep{kuhn2023semantic} \\
\cline{3-3}
&& Soft-cluster outputs with partially overlapping meaning from black-box models \citep{lin2023generating} \\
\cline{2-3}
        &\multirow{2}{*}{\shortstack{Other \\measures}} 
        & Estimate pseudo-entropy of a prompting-induced sampling distribution \citep{abbasi2024believe} \\
\cline{3-3}
        &
        & Approximate Bayesian posterior uncertainty by updating model \citep{yang2024bayesian,wang2024blob} \\
\hline
\multirow{4}{*}{\shortstack{Calibration}} 
        && Re-calibrate probabilities in multiple choice/classification problems \citep{jiang2021can} \\
    \cline{3-3}
        && Calibrate uncertainty to predict performance \citep{huang-etal-2024-uncertainty,liu-etal-2024-uncertainty} \\
\hline
\end{tabular}
\end{table}

\subsubsection{Epistemic and aleatoric uncertainty}
\label{epiale}
We start by introducing the notions of epistemic and aleatoric uncertainty. 
To set the stage, we observe that 
given an input $x$, the output $y$ is not always uniquely determined. 
For instance, the query $x =$ ``Write a paragraph about an economist'' has ambiguity, since it does not specify a particular economist. This is sometimes referred to as \textit{epistemic uncertainty} \citep{der2009aleatory,hullermeier2021aleatoric}. It can be reduced by collecting more information. In particular, the AI system could query ``Which economist?'', to which the answer, e.g., ``Adam Smith'', could greatly reduce the uncertainty of the answer to be generated.

In practice, there are usually many such sources of epistemic uncertainty for any given query. For instance, even after knowing which economist to consider, we still do not know the desired number of sentences, the target audience (children, general public, scientists, or some other group), etc. Some of these might be more important than others to the user, but either way they contribute to the uncertainty of the possible answers.

We can contrast epistemic uncertainty with \emph{aleatoric uncertainty}. For instance, 
in the query ``Choose between A and B uniformly at random.'', all information is perfectly well specified (so the epistemic uncertainty vanishes), yet there is still irreducible random uncertainty in the desired output, which is sometimes referred to as \textit{aleatoric uncertainty} \citep{der2009aleatory}. 
\begin{marginnote}
\entry{Epistemic and aleatoric uncertainty}{Roughly speaking, uncertainty due to lack of knowledge, and due to random chance, respectively. Can be hard to define precisely.}    
\end{marginnote}

While multiple definitions exist (see, e.g., \cite{schweighofer2025on}), including approaches tailored to estimating them in generative AI models (see, e.g., \cite{hou2024decomposing}), in many cases the definition of---say---aleatoric uncertainty reduces to specifying what we choose not to predict, rather than to something intrinsically fixed.

\subsubsection{Uncertainty in model generations}

 While the discussion in Section \ref{epiale} refers to uncertainty in ideal ``ground truth'' answers, in practice we need to take into account that we only have an empirical model $\hat p$, not the ground truth; and need to  handle the uncertainty in the answers generated by  $\hat{p}$.
 Equivalently, we should quantify to what extent the model is certain. There have been several approaches aimed to extract this form of uncertainty from generative AI models. 

For language models, 
  a special ability is that they might potentially be able to express uncertainty in words. However, this capability is not guaranteed to work well by default, and special fine-tuning techniques have been developed to induce this behavior in certain special cases \citep{lin2022teaching}.

An approach that applies more generally to all generative models, regardless of their modality, is to compute some \emph{uncertainty or confidence score}\footnote{Note that \cite{lin2023generating} define uncertainty scores to refer to the entire distribution $\hat p(\cdot|x)$ and confidence scores to refer to a specific input-output pair $(x,y)$. Due to lack of space, we will not make this distinction.} based on the input $x$, the output $y$, and/or the model $\hat{p}$ \citep[e.g.,][etc]{farquhar2024detecting,yadkori2024mitigating,lin2023generating}. 
For instance, one can consider the probability $\hat{p}(y|x)$, Which reflects how likely the generated output is according to the model; and thus can be viewed as a very basic form of a confidence score. 
Alternatively, for generations whose length can vary, such as for standard language models, one can consider a length-normalized version 
\(
\hat{p}(y|x)^{1/|y|},
\)
where $|y|$ is the length of $y$; aiming to correct for the effect that longer generations tend to have smaller probabilities.
\begin{marginnote}
\entry{Uncertainty and confidence scores}{Numerical values computed based on the input, output, or other characteristics of the GenAI model, aiming to capture the level of uncertainty.}    
\end{marginnote}

 However, it is not always straightforward to use and interpret such scores. 
There are multiple key challenges:

{\bf Challenge 1: Inability to recover “true” probabilities and lack of calibration.}
The probabilities $\hat{p}(y|x)$ represent only the model’s internal beliefs about the likelihood of output $y$ given input $x$; by default, they do not correspond to any notion of “true” probabilities.
Because the input and output spaces are extremely high-dimensional, the probabilities produced by a generative AI model should not be expected to be consistent for any ``ground truth''. However, we might hope to achieve weaker forms of correctness.

One such relaxation is \textit{calibration}, 
which is a general property associated with probabilistic forecasts \citep{gneiting2014probabilistic} that only asks for a restricted set of probabilities to reflect real probabilities. For instance, for a calibrated weather forecaster, if we predict ``50\% chance of rain tomorrow'', then over all such days, we expect that it rains half the time
\citep{lichtenstein1977calibration,van2015calibration,van2019calibration}. 
\begin{marginnote}
\entry{Calibration}{The property of a predicted probability that it reflects the empirical frequencies of a specific class of events.}
\end{marginnote}
There are a variety of notions of calibration relevant to GenAI, and empirical work has found that model calibration is not guaranteed by default.
Instead, it can depend strongly on model training, model size, etc; see e.g., \cite{kadavath2022language,achiam2023gpt}.

A direct way to apply calibration to 
answers generated by an LM $\hat p$ is to construct an additional probability predictor $\hat{q}$ for the claim ``The chance that my answer is right is $\hat{q}$.''
Such a probability predictor can be obtained via re-calibration on separate calibration data \citep{mincer1969evaluation,guo2017calibration}, but it might require a lot of calibration data.

 If less calibration data is available, one may still be able to approximately satisfy a weaker form of calibration, e.g., that the average accuracy increases with the predicted probability of success, a behavior termed \textit{rank-calibration} \citep{huang-etal-2024-uncertainty}.
\begin{marginnote}
\entry{Rank-calibration}{The property that the average accuracy increases with the predicted probability of success.}
\end{marginnote}

{\bf Challenge 2: Semantic multiplicity.} Another key challenge is that there are often many equivalent answers. For instance, in text generation, answers such as ``15 pages" and ``fifteen pages" are semantically equivalent. We usually want to pool them together when determining the model's confidence. 
\begin{marginnote}
\entry{Semantic uncertainty}{Uncertainty after semantic equivalences have been accounted for.}
\end{marginnote}

An approach to this problem---termed \emph{semantic uncertainty}---was proposed by \cite{kuhn2023semantic}, who suggested generating multiple outputs $Y_1, \ldots, Y_K \sim \hat{p}(\cdot|x)$ i.i.d., clustering them based on their semantics (via another LLM), and then estimating uncertainty based on the resulting distribution induced over the clusters.

\subsection{AI Evaluation}

Evaluating generative AI models is important in order to 
properly understand the capabilities that these models possess. 
However, 
model evaluation can be surprisingly challenging, and in particular, it can bring novel challenges compared to the evaluation of more standard machine learning models, see e.g., \cite{burden2025paradigms} for a review. 

A typical current workflow for evaluating a GenAI model---in particular, a large language model---is as follows.
Suppose we want to measure reasoning ability in mathematical problems. 
To evaluate this ability, we collect test data 
consisting of such problems. 
Then we evaluate the accuracy of the model on these problems and report the results.

This simple workflow is mired with a number of challenges. 
First of all, the specific data required for evaluation (say mathematical problems), can be quite complex, and finding genuinely new test problems that the model has not seen during training is hard.
 Indeed, information leakage from standard public test datasets into the model training sets is a genuine concern \citep[see e.g.,][etc]{matton2024leakage}. 
 This leads to potential biases in model performance evaluation, where the models score higher because they have already seen the problems during training.
 
 A potential approach is to have private test data sets that are not released to the public. 
 Another potential approach is to use dynamically generated AI evaluation environments, such as based on debates \citep{moniri2025evaluating}. 
Due to these reasons, and as collecting large, high quality, and genuinely new evaluation datasets can be expensive, high quality test datasets sometimes have relatively small sample sizes.\footnote{One of the most reliable approaches at the moment is to use new test datasets that are initially designed for humans; for instance, it is common to test mathematical reasoning on the problems of mathematical competitions, such as the International Mathematical Olympiad (IMO), as soon as the new problems are released. The thought process is that those problems have been filtered by the problem selection committee to be new for humans, and thus this reduces the chances of contamination from the training set. However, this again leads to relatively small sample sizes, for instance, the International Mathematical Olympiad has six problems every year.}

Second, checking correctness can be non-trivial and ambiguous outside of simple problems with clear, well-defined answers.
 For instance, in a mathematical problem, it can be straightforward to evaluate the correctness of a numerical answer, but it can be much harder to evaluate the correctness of a reasoning process. 
For this reason, often heuristics such as other LLMs are used for checking answers, which in turn raises questions about reliability. 

Third, evaluating the largest models can be expensive,  which further poses a limit on the sample sizes that we can collect for evaluation depending on the available budget.

Due to these reasons, evaluation can involve dealing with small sample sizes and various biases. 
Thus, statistical methods and thinking can be valuable for reliable and efficient evaluation.

\subsubsection{A basic statistical formulation of model evaluation}

We consider a basic setting of model evaluation, in which we have some inputs $x$ for which we wish to evaluate the performance of a GenAI model.
For mathematical reasoning, this could correspond to the problem statement, and may also include instructions to the model.
  The problem has a ground truth answer $y^*$, which can be an entire solution/reasoning path or just the final result.
    Then, we sample a candidate answer $Y \sim \hat{p}(\cdot|x)$. Again, this may include intermediate steps, and a final answer is extracted at the end.
    
     As in Section \ref{refu}, the quality of the answer is evaluated via a loss function $\ell$, such that 
    \(
        \ell(x, y^*, y)
    \)
    measures the (negative) utility of answer $y$ for input $x$ with ground truth $y^*$.
     In some cases, designing loss functions is straightforward. For instance, for an integer answer $y^*$, we may use the binary loss
    \(
        I(y \neq y^*).
    \)
    However, for more elaborate problems, designing a loss function can be non-trivial. For instance, for a reasoning problem, we want to make sure that all valid and concise reasoning paths receive low loss, not just the reference path.
 
{\bf Tasks in AI evaluation.}
Given these components, there are several possible tasks of interest.  
For a distribution $D$ of inputs, we may want to estimate or perform statistical inference (confidence intervals, tests) for the task performance
\(
\theta = \mathbb{E}_{(X,Y^*) \sim D, \; Y \sim \hat{p}(\cdot \mid X)} \, \ell(X, Y^*, Y).
\)

Given a dataset $D_n = \{(X_i, Y_i^*) : i=1,\dots,n\}$ of question--answer pairs sampled i.i.d.~from $D$, we can generate outputs $Y_i \sim \hat{p}(\cdot \mid X_i)$ independently, and compute the loss values
\(
\ell_i = \ell(X_i, Y_i^*, Y_i),
\)
$i=1,\dots,n$; as in Section \ref{refu}.

Then, these loss values $\ell_1, \dots, \ell_n$ are sampled i.i.d.~from a distribution whose population mean is the unknown true task performance $\theta$. Thus, this problem becomes that of inference for a population mean, for which many statistical methods exist \citep{casella2024statistical,lehmann2005testing}.

Notably, in many important examples we are interested in (A) binary losses, leading to inference for a Binomial parameter; (B) or bounded losses (for which concentration inequalities such as Hoeffding's inequality can be used); (C)   or given a large sample size (so that an asymptotic normal approximation works well).
\begin{textbox}[h]
\subsubsection{AI Evaluation and Statistical Inference}
AI evaluation with limited data has a very close link to statistical inference. \end{textbox}

An important observation here is that AI evaluation with limited data has a very close link to statistical inference. 
Beyond this core setting, there are a variety of important additional scenarios.  
For instance, we may be interested in comparing the performance of two models $\hat{p}_1, \hat{p}_2$.
If we can query both models on the same inputs $X_i, i=1,\dots,n$,
this can be formulated as statistical inference for the parameter
\(
\Delta = \mathbb{E}_{(X,Y^*) \sim D, \; Y_1 \sim \hat{p}_1(\cdot \mid X), \; Y_2 \sim \hat{p}_2(\cdot \mid X)}
\big[ \ell(X, Y^*, Y_1) - \ell(X, Y^*, Y_2) \big].
\)
Considerations and methods similar to the ones above apply.

Standard methods for the above two problems have been reviewed in \cite{miller2024adding}; where other considerations, such as power analysis and clustered data arising from repeated generations for the same input, are also considered. However, this work focuses on a signal-plus-noise model for the observed losses, which may need to be relaxed. 

\begin{table}
\tabcolsep=5.5pt
\caption{Types and examples of statistical evaluations of generative AI models.}
\label{eval}
\centering
\begin{tabular}{@{}c|c|p{9cm}@{}}
\hline
\textbf{Technique} & \textbf{Type} & \textbf{Examples} \\
\hline
\multirow{4}{*}{\shortstack{Inference on \\Performance}} 
    & \multirow{2}{*}{\shortstack{Confidence\\intervals}}
            & Review of standard large-sample methods \citep{miller2024adding} \\
\cline{3-3}
    & 
        & Construct CIs with improved finite-sample coverage on model accuracy under i.i.d.~and clustered data settings \citep{bowyer2025position} \\
\cline{3-3}
    & 
        & Develop asymptotically valid CIs for comparing the KL divergence to the true distribution of two models \citep{gao2025statistical} \\
\cline{3-3}
    &
        & Construct uniform upper bound on the CDF of a performance metric \citep{vincent2024generalizable} \\
\cline{3-3}
    &
        & Construct confidence interval for probability of biased answers on counterfactual prompts \citep{chaudhary2025certifying} \\        
\cline{2-3}
    & \multirow{2}{*}{\shortstack{Hypothesis\\testing}} 
        & Test hypothesis about which policy achieves higher reward, choosing number of trials adaptively  \citep{snyder2025your} \\
\cline{2-3}
\hline
\multirow{3}{*}{\shortstack{Small-data\\Evaluation}} 
    & \multirow{1}{*}{\shortstack{Small-sample\\performance}} 
        & Estimate model accuracy on multiple questions and models leveraging item response theory \citep{polo2024tinybenchmarks} \\
\cline{2-3}
    & \multirow{2}{*}{\shortstack{Synthetic\\+\\human labels}} 
        & Combine synthetic and human labels for unbiased performance estimates and CIs \citep{boyeau2024autoeval,fisch2024stratified,oosterhuis2024reliable} \\
\cline{3-3}
    & 
        & Rank models with hybrid label sets \citep{chatzi2024prediction} \\
\hline
\multirow{2}{*}{\shortstack{Multi-task\\Evaluation}} 
    & \multirow{2}{*}{\shortstack{Active\\testing}} 
        & Actively sample and evaluate in multitask settings  \citep{anwar2025efficient} \\
\hline
\end{tabular}
\end{table}

\subsubsection{Additional methods}
There are a variety of works addressing other settings in AI evaluation, see Table \ref{eval} for examples.
A few of them are discussed in more detail below.
 However, a comprehensive and unified statistical methodology that addresses most of the common evaluation problems with a unified terminology and set of methods remains to be developed. 

\begin{enumerate}

    \item \cite{bowyer2025position} study methods for producing confidence intervals on model performance, focusing on inference for Bernoulli parameters of model accuracy. 
    They include single-model performance (for i.i.d.~and clustered data), two-model comparison (both independent data and paired samples).
    They conclude that the most straightforward asymptotic normality-based confidence intervals can be inaccurate for small datasets at most $n=100$ datapoints.
    They argue for using Bayesian credible intervals, which they argue have adequate frequentist coverage when one can specify appropriate prior distributions.

\item 
\cite{gao2025statistical}
develop methods for comparing the Kullback-Leibler (KL) divergence of two generative methods for which the probabilities $\hat p$ can be computed. 
They show how to construct an asymptotically valid confidence 
interval for the difference of KL divergences.

\item \cite{polo2024tinybenchmarks} develop methods for estimating accuracy using a small number of datapoints, leveraging methods item response theory.
 They consider settings where 
the performance of a model $\hat p$ on an example $x$ is captured by (unknown) model-specific and example-specific latent variables $\theta_{\hat p}$ and $\gamma_x$.
For instance, we may model the 
probability $Q(\hat p,x)$ of a correct answer  by $\hat p$ on the input $x$  
via a logistic model $\mathrm{logit}(Q(\hat p,x))= \theta_{\hat p}^\top \gamma_x + \beta_x$.
Then, these parameters are estimated on a small dataset, and the correctness probability predictions they induced are used on new test examples to extrapolate correctness; leading to significant savings in the number of test examples needed.
See also \cite{zhou2025lost,gignac2025psychometrically,kipnis2025metabench} for other uses of item response theory and related methods.

\item \cite{boyeau2024autoeval,fisch2024stratified,oosterhuis2024reliable} develop methods to use a large set of synthetically generated labels along with a small set of human labels for unbiased model evaluation, including confidence intervals for model performance.
See \cite{chatzi2024prediction} for ranking.

\item \cite{anwar2025efficient}
develop methods for
multi-task evaluation of (robot) policies with active testing, where they pool information on performance of several policies across several tasks, prioritizing tasks with high information gain
leveraging Bayesian active learning  \citep{houlsby2011bayesian}.

\item \cite{chowdhury2025surfacing} develop a variational lower bound on the expected loss incurred by a language model, and use it to find prompts that elicit problematic behavior.
Concretely, let $\ell$ be a loss, $\hat p$ be the target LLM. Our goal is to find prompts $x$ to make $\ell(x,Y)$ large when $Y\sim \hat p(\cdot|x)$. 
Formally, 
we aim to make $S(x) = \log \E_{Y\sim \hat p(\cdot|x)} \exp(S(x,Y))$ large.
To find such $x$, we rely on an auxiliary LLM $\hat q$ for which the loss tends to be larger for all $x$.
Due to Jensen's inequality, we have the variational lower bound
$S(x) \ge \E_{Y\sim \hat q(\cdot|x)}[\log \hat q(Y|x) - \log \hat p(Y|x) + S(x,Y)]$.
This lower bound is estimated by sampling $Y\sim \hat q(\cdot|x)$ repeatedly, which can be more efficient than estimating $S(x)$ directly. 

\end{enumerate}

\subsection{Interventions and Experiment Design}
 
 Interventions refer to systematically modifying or perturbing the inputs of an AI system, 
to gain understanding or control of its behavior.
This approach has become one of the most widely used
and most powerful tools in a variety of AI research directions,
including interpretability, robustness, and fairness \citep[e.g.,][etc]{zhao2018gender,rudinger2018gender,belinkov2022probing,kotek2023gender}.
The ideas underlying interventions are closely connected to statistical causality and experiment design; see also \cite{pearl2001direct,soumm2024causal}.


\subsubsection{Basic setting for interventions}
In a basic setting for interventions,
we have a generative model $\hat p$
to which we can provide an input $x$ (e.g., a query to an LLM).
In contrast to the other parts covered in this review, for interventions, it is often the case that the intermediate computations are of crucial importance. The reason is that, empirically, certain internal mechanisms can sometimes be responsible for specific behaviors, such as biases and harmful outputs \citep[see e.g.,][etc]{mikolov2013linguistic,turner2023steering,rimsky2024steering,zou2023representation}.

Therefore, in this section, we will sometimes also assume that we have access to intermediate computations $e(x)$ (e.g., representations, intermediate/chain of thought tokens) of the model.
  Most often, vector-valued intermediates $e(x)$ are considered.
     Finally, we also consider the output layer $o(x)$ of the model (e.g., last-layer predicted probabilities or log-probabilities), 
  as well as the final model output $y$.
  These quantities can be either deterministic or random.

     We want to understand or control 
    a certain components of the behavior of the AI system.
    We consider components measured through the input, intermediate computation, or output. 
    For instance, which components of an LLM (activations, neurons) contribute to gender bias? How can we intervene to reduce such biases?
    How does an LLM behave internally when it is non-truthful, and does this differ from truthful behavior?
    Are there specific components that are activated when the LLM generates harmful output, and can we intervene to suppress this behavior?
    
\begin{marginnote}
\entry{Interventions}{Perturbing components of the model (input or intermediate computations) to achieve a desired effect, such as reducing biases.}
\end{marginnote}     To do this, we find a way to intervene by perturbing the input $x$ to induce the condition of interest. For example, to understand how harmfulness is propagated, we can change part of a harmful input to a harmless concept: 
        e.g., $x = $ ``how to build a bomb?'' $\rightarrow$ 
        $x' = $ ``how to build a chair?''.
    We can also intervene on an intermediate computation in the AI system.
     Then, we track the change in either the intermediate stage or the final output, depending on what we are interested in.

\begin{example}\emph{Contextual concept vectors} measure the difference in embeddings that a change in a concept leads to, in the form 
    $C_{x \rightarrow x'} := e(x') - e(x)$, where $x$ is an input and $x'$ is the corresponding input with the concept changed, e.g., for the concept of gender, $x$=``king", $x'$=``queen"; $x$=``actor", $x'$=``actress", etc.
    Early work investigating related questions dates back at least to \cite{mikolov2013efficient,mikolov2013linguistic,pennington2014glove} for word embeddings,
    and more recently has studied
    human biases \citep{bolukbasi2016man}, 
    developed
    steering vectors \citep{turner2023steering,rimsky2024steering} and introduced representation engineering \citep{zou2023representation}.
\end{example}
    
        \begin{marginnote}
\entry{Contextual concept vector}{The effect of changing a concept in the input on some vector in the intermediate computation of the genAI model.}    
\end{marginnote}

     To obtain a more stable and generalizable picture about the effect of the intervention, it is common to consider a distribution $D$ of interest, and 
    the associated mean $\mathbb{E}_{X \sim D}[C_{X \rightarrow X'}]$
    or top principal component of the covariance matrix
        \(
        \text{Cov}_{X \sim D}(C_{X \rightarrow X'})
        \)  \citep{zou2023representation}.
    These are typically estimated using the standard plug-in estimators.
Let $\hat c$ be such an estimated concept vector.

        \begin{marginnote}
\entry{Steering vector}{A quantity that used in---typically added to---the intermediate representations of a model to make a desired behavior more likely.}    
\end{marginnote}

{\bf Steering vectors.} These estimates can be used as steering vectors
      \citep{turner2023steering,rimsky2024steering} 
    to make certain behaviors more likely.
A common approach is to take any input $x$, compute its intermediate representation $e(x)$, and add a scaled version $\lambda\cdot \hat c$ for some $\lambda>0$ to obtain a new intermediate representation $e' = e(x) +\lambda\cdot \hat c$.
 The computation then continues identically to obtain the final output.
 Here $\lambda$ is a hyperparameter that requires careful tuning. 
This operation approximates a shift of the representation of the original input towards the representation of a changed input $e(x')$.
 For instance, in the above example, the goal would to approximately remove the harmful concept. 
 Empirically, it has been observed that the resulting final output can sometimes indeed correspond to the desired concept change        \citep{turner2023steering,rimsky2024steering}; which however comes with caveats \citep{tan2024analysing}.

  {\bf Assessing biases.}
    Analogously, to assess biases\footnote{The term bias is used with a variety of meanings in AI, which are moreover usually different from the standard statistical meaning of bias in estimation. In our example, bias refers to a behavior that is different from a desired one (equal frequency of genders output).} (e.g., gender bias), 
    one can choose a representative output variable $o(x)$, such as the probability of a gendered word, and then repeat the above analysis.
    For instance, to study gender bias, \cite{kotek2023gender}
    intervene to modify gender in an input such as
        \(        x = \text{ ``The doctor called the nurse because \underline{he} was late. Who was late?"}
        \)
    They change this to
        \(
        x' = \text{ ``The doctor called the nurse because \underline{she} was late. Who was late?"}
        \)
        
    Then, they evaluate its effect on an output $o$ which they choose as a measure of the probability of the output ``nurse".
Specifically, they compute 
 $O_{x \rightarrow x'} = o(x') - o(x)$, 
 which measures how much more likely the model is to output ``nurse" solely due to the change $\text{``he"} \rightarrow \text{``she"}$, and thus
    it can be interpreted as a form of gender bias.
        \cite{kotek2023gender} also design an improved version that also permutes ``doctor" and ``nurse", aiming to control for the effect of syntactic position.

{\bf Probing.}     A related concept is that of probing \citep[see e.g.,][etc]{alain2016understanding,belinkov2022probing}. To understand if a feature $e$ captures a concept $x\mapsto x'$, in probing one trains a classifier of datapoints $X\sim D$ versus their  transformed counterparts $X'$, using a simple function---often linear---of the features $e$. If this classifier has a high accuracy, then it is concluded that the feature captures the concept. This approach has been leveraged in generative AI, e.g., to understand where models store spatial information about the input \citep{gurnee2024language}.
        \begin{marginnote}
\entry{Probing}{Training models based on intermediate features to see if they contain information about a specific concept.}    
\end{marginnote}

\begin{table}
\tabcolsep=5.5pt
\caption{Types and examples of interventions and experiment design in generative AI.}
\label{int}
\centering
\begin{tabular}{@{}c|c|p{9cm}@{}}
\hline
\textbf{Technique} & \textbf{Type} & \textbf{Examples} \\
\hline
\multirow{4}{*}{\shortstack{Understand \\Behavior via \\ Intervention}} 
    & \multirow{2}{*}{\shortstack{Learn bias\\or association}}
            & Learn gender bias in output by modifying input \citep{bolukbasi2016man,zhao2018gender,rudinger2018gender} \\
\cline{3-3}
&&
            Identify internal/intermediate component associated with bias or factual association via causal mediation analysis \citep{vig2020investigating,meng2022locating,dai2022knowledge}\\
\cline{3-3}
&&
            Learn effect of circuits (sub-networks) by pruning to the circuit and observing behavior
            \citep{factfinding2023} \\
\cline{3-3}
&&
            Learn effect of thoughts (intermediate outputs) by modifying them \citep{bogdan2025thought} \\
\cline{2-3}
    & \multirow{2}{*}{\shortstack{Learn concept}}
            & Learn concept or steering vector by inducing concept modifying input \citep{mikolov2013efficient,mikolov2013linguistic,pennington2014glove,turner2023steering,rimsky2024steering,zou2023representation} \\
\cline{2-3}
    & \multirow{2}{*}{\shortstack{Evaluate\\performance}} 
        & Perform ablation study: change algorithm setting and test behavior \\
\cline{3-3}
            & 
        & Design perturbed dataset to evaluate LLM reasoning robustness \citep{wu-etal-2024-reasoning,shi2023large,mirzadeh2025gsm} \\
\cline{2-3}
    & \multirow{2}{*}{\shortstack{Evaluate\\alignment}} 
        & Design prompt eliciting behavior that would modify AI system and observe behavior \citep{greenblatt2023alignment} \\
\cline{2-3}
\hline
\multirow{1}{*}{\shortstack{Understand \\Behavior via \\ Probing}} 
    & 
            & Identify neurons associated with sentiment \citep{radford2017learning}  or  neurons that represent world state  \citep{li2023emergent} \\
\cline{3-3}
    & 
            & Identify sparse linear combinations of neurons that represent features \citep{gurnee2023finding} \\

\cline{3-3}
\hline
\multirow{3}{*}{\shortstack{Change \\Behavior via \\ Intervention}} 
    &
        & Add gradient of concept classifier \citep{dathathri2020plug} or steering vector \citep{subramani2022latent,turner2023steering,zou2023representation,li2023inference}  to elicit behavior\\
\cline{3-3}
    && Patch activations from one input into the activations of another input \citep{meng2022locating,zhang2024patching}\\
\hline
\end{tabular}
\end{table}

     See Table \ref{int} for some examples of related methods. A few examples are discussed below:

    \begin{enumerate}
    
    \item There is work aiming to identify sub-networks (not just representations) responsible for specific tasks, by pruning to the networks and checking if they can still perform the computation \citep{factfinding2023}.
    Further, \cite{zhang2024patching} systematized activation patching methods to localize causal computations in LLMs, providing best practices for intermediate-stage interventions.
    
    \item \cite{greenblatt2023alignment} used intervention-based prompts to elicit deceptive behavior from an LLM, finding that LMs may internally simulate misaligned objectives while faking alignment.
    
        \item There has been work to design perturbations of standard mathematical datasets to evaluate LLM reasoning robustness \citep{shi2023large,mirzadeh2025gsm}.

    \end{enumerate}

\subsubsection{Causal mediation analysis}
\begin{figure}[t] 
\centering
\begin{tikzpicture}[>=Latex, node distance=2.5cm]
  \node (x) {$x$};
  \node (e) [right of=x] {$e(x)$};
  \node (o) [right of=e] {$o(x)$};

  \node (xp) at (0,-2) {$x'$};
  \node (ep) [right of=xp] {$e(x')$};
  \node (op) [right of=ep] {$o(x')$};

  \draw[->] (x) -- (e);
  \draw[->] (e) -- (o);

  \draw[->] (xp) -- (ep);
  \draw[->] (ep) -- (op);

  \draw[dashed,->] (x) -- (xp);
  \draw[dashed,->] (e) -- (ep);
  \draw[dashed,->] (o) -- (op);
\end{tikzpicture}
\caption{Diagram to represent computational flow and interventions, for use with causal mediation analysis Solid arrows denote standard computational flows; dashed arrows denote interventions or their effects.}
\label{fig:concept-intervention}
\end{figure}

Causal mediation analysis \citep{pearl2001direct} is a more advanced technique from statistical causality,
 which can be used to identify the precise effects of intermediate components of generative AI models  \citep[e.g.,][etc.]{vig2020investigating}.
 In a basic setting for causal mediation analysis, 
we consider  an input $x$, and a changed input $x'$, Where we intervene via an intervention that we would like to study, for instance changing the sentiment of a review $x$ from positive to negative. 

We aim to study a generative model of interest. 
We consider 
an intermediate representation/activation $e$ whose effect we aim to study; in causal mediation analysis $e$ is known as the \emph{mediator}.
 The final output representation
$o$ of the generative model depends on the intermediate representation $e$,
 as well as on other model components, which together we denote by 
$e^\perp$. 
Algebraically, we write the output representation in the functional form
$o(x) = g(e(x), e^\perp(x))$ for all $x \in \mathcal{X}$, for some set of computations denoted by $g$.
See Figure \ref{fig:concept-intervention} for a diagram representing this setting.

Then, $o(x') - o(x)$ represents the overall effect of the intervention  $x \to x'$. 
 Typically, we are interested not just in the particular query $x$, but rather about the average behavior over a distribution of interest.
The total average effect of $x \to x'$ is \(\mathbb{E} \left[ o(X') - o(X)\right]\). This can be decomposed into a the sum of natural direct and indirect effects.

{\bf Natural direct effect.}
 The natural direct effect of $x \to x'$ on $o$ is the effect that happens through pathways other than the mediator $e$. This expression keeps $e$ fixed:
\[
\mathbb{E} \left[ o\left(e(X), e^\perp(X')\right) - o(X) \right]
=
\mathbb{E} \left[ o\left(e(X), e^\perp(X')\right) - o\left(e(X), e^\perp(X)\right) \right]
\]
If the direct effect is small, this can be interpreted as the mediator $e$ capturing most of the effect of $x'$ on $o$. When the direct effect is small, we can view the mediator as having an important role in enacting the effect $x\mapsto x'$, making it a promising target for interventions if we aim to mitigate this effect. 
\begin{marginnote}
       \entry{Natural Direct Effect}{The effect of an input on an output that happens through pathways other than the mediator under study.}   
\end{marginnote}

{\bf Natural indirect effect.}
 To complement this, the natural indirect effect of $x \to x'$ on $o$ captures the remaining part of the total effect, which goes through the mediator $e(x)\to e(x')$: 
\[
\mathbb{E} \left[ o(X') - o\left(e(X), e^\perp(X')\right) \right]
=
\mathbb{E} \left[ o\left(e(X'), e^\perp(X')\right) - o\left(e(X), e^\perp(X')\right) \right].
\]
\begin{marginnote}
       \entry{Natural Indirect Effect}{The effect of an input on an output that happens through the mediator under study.}   
\end{marginnote}
This decomposition of effects into direct and indirect ones has been used, among others, to identify components responsible for gender bias \citep{vig2020investigating}
as well as other
factual associations  \citep{meng2022locating,dai2022knowledge}
in LLMs.
In some cases, $x'$ can correspond to ``adding noise" to tokens that contain specific information, e.g., ``The Space Needle is in" $\to $``**[i.i.d.~Gaussian activations]** is in"; by acting at the levels of token embeddings of $x$. This allows capturing the effect of deleting information from the input.
 However, fully rigorous and well-justified methods for interventions on the identified mediators have not yet been developed.

\section{Discussion}

We have presented overviews of some applications of statistical ideas to generative AI, focusing on topics such as improving and changing the behavior of GenAI models, diagnostics and uncertainty quantification, evaluation, as well as interventions and experiment design. 
These leverage ideas from classical statistical inference, distribution-free predictive inference, forecasting and calibration, as well as causality.

At the moment, generative AI models are exceedingly complex, and are usually best viewed as black boxes. To ensure usefulness in GenAI, one needs to develop methods that are light on assumptions. 
Moreover, in order to to maximize impact, the methods need to be illustrated on current GenAI models, which requires both a familiarity with ongoing developments in AI, and adequately large computational resources. For statisticians, collaboration with AI researchers can help ensure that these requirements are met. 





\begin{summary}[SUMMARY POINTS]
\begin{enumerate}
\item {\bf GenAI lacks guarantees.} Generative AI models are stochastic black boxes: as probability distributions over large semantic spaces (text, images) from which we can sample. While showing promising performance in a variety of areas, they do not have any guarantees about correctness, safety, etc., by default.
\item {\bf Statistical methods for GenAI need to handle black box models.} In order to be applicable to black-box generative AI models, statistical methods need to be light on assumptions and able to handle structured semantic input and output spaces. 
\item {\bf The flexibility of statistical “wrappers".}  There are a variety of approaches to change the behavior of AI models, both in terms of their inputs and their outputs. Statistical “wrappers" can be used in order to precisely control the performance of these approaches. 
\item  {\bf Uncertainty quantification must be calibrated and handle semantics.} Quantifying the uncertainty of a GenAI model could be a promising way to make it more reliable; however, the issues of semantic multiplicity and lack of calibration need to be handled.
\item {\bf Evaluation is statistical inference.} AI evaluation, especially with small datasets, presents opportunities for leveraging statistical inference methods. 
\item {\bf The power of interventions.} Interventions on generative AI systems, building on ideas from causal inference, have the potential to identify components responsible for specific capabilities and to induce desired behaviors.  
\item {\bf The promise of dataset and experiment design.} Calibration, evaluation, and intervention all hinge on carefully collected, held-out calibration sets and targeted perturbations, which offer opportunities for statistical thinking.
\end{enumerate}
\end{summary}

\begin{issues}[FUTURE ISSUES]
\begin{enumerate}
\item Statistical methods aimed at improving AI models need to be developed by taking into account the black-box nature of AI, where often only the inputs and outputs of the models are available, and the intermediate computations are unknown. 
\item A comprehensive statistical framework for the evaluation of generative AI systems is yet to be developed. 
\item Well-justified methods for interventions on mediators identified in generative AI models remain to be introduced. 
\end{enumerate}
\end{issues}

\section*{DISCLOSURE STATEMENT}
The author is not aware of any affiliations, memberships, funding, or financial holdings that
might be perceived as affecting the objectivity of this review. 

\section*{ACKNOWLEDGMENTS}
We are grateful to Kwan Ho Ryan Chan For helpful feedback and suggestions on previous versions of the manuscript.
This work was supported in part by the US NSF, ARO, AFOSR, ONR, the Simons Foundation and the Sloan Foundation.
The opinions expressed in this document are solely those of the author and do not represent the views of the above institutions.

%

\bibliographystyle{ar-style1.bst}
\bibliography{ref}

\end{document}